\documentclass[sigconf,nonacm]{acmart}

\usepackage{listings}
\usepackage{xcolor}
\usepackage{adjustbox}
\usepackage{float}
\usepackage{color,soul}
\usepackage{aliascnt}
\usepackage{amsmath}
\usepackage{textgreek}
\usepackage{caption}
\usepackage{tikz-cd}
\usepackage{graphicx}
\usepackage{multirow}
\usepackage{hyperref}
\usepackage{academicons}
\usepackage{booktabs}
\restylefloat{table}
\usepackage{comment}
\usepackage{subcaption}
\usepackage{float}

\newaliascnt{eqfloat}{equation}
\newfloat{eqfloat}{h}{eqflts}
\floatname{eqfloat}{Equation}
\newcommand*{\ORGeqfloat}{}
\let\ORGeqfloat\eqfloat
\def\eqfloat{%
  \let\ORIGINALcaption\caption
  \def\caption{%
    \addtocounter{equation}{-1}%
    \ORIGINALcaption
  }%
  \ORGeqfloat
}

\colorlet{punct}{red!60!black}
\definecolor{background}{HTML}{EEEEEE}
\definecolor{delim}{RGB}{20,105,176}
\colorlet{numb}{magenta!60!black}

\lstdefinelanguage{json}{
    basicstyle=\normalfont\ttfamily,
    numbers=left,
    numberstyle=\scriptsize,
    stepnumber=1,
    numbersep=8pt,
    showstringspaces=false,
    frame=lines,
    literate=
     *{0}{{{\color{numb}0}}}{1}
      {1}{{{\color{numb}1}}}{1}
      {2}{{{\color{numb}2}}}{1}
      {3}{{{\color{numb}3}}}{1}
      {4}{{{\color{numb}4}}}{1}a
      {5}{{{\color{numb}5}}}{1}
      {6}{{{\color{numb}6}}}{1}
      {7}{{{\color{numb}7}}}{1}
      {8}{{{\color{numb}8}}}{1}
      {9}{{{\color{numb}9}}}{1}
      {:}{{{\color{punct}{:}}}}{1}
      {,}{{{\color{punct}{,}}}}{1}
      {\{}{{{\color{delim}{\{}}}}{1}
      {\}}{{{\color{delim}{\}}}}}{1}
      {[}{{{\color{delim}{[}}}}{1}
      {]}{{{\color{delim}{]}}}}{1},
}

\AtBeginDocument{%
  }

\begin{document}

%%
%% The "title" command has an optional parameter,
%% allowing the author to define a "short title" to be used in page headers.
\title[Generating synthetic financial time series]{Fiaingen: A financial time series generative method matching real-world data quality}

%%
%% The "author" command and its associated commands are used to define
%% the authors and their affiliations.
%% Of note is the shared affiliation of the first two authors, and the
%% "authornote" and "authornotemark" commands
%% used to denote shared contribution to the research.
\author{Jo\v{z}e M. Ro\v{z}anec}
\email{joze.rozanec@utwente.nl/joze.rozanec@ijs.si}
\orcid{0000-0002-3665-639X}
\affiliation{%
  \institution{University of Twente/Jo\v{z}ef Stefan Institute}
  \city{Enschede/Ljubljana}
  \country{The Netherlands/Slovenia}
}

\author{Tina \v{Z}ezlin}
\email{tina.zezlin@gmail.com}
\affiliation{%
  \institution{Jo\v{z}ef Stefan Institute}
  \city{Ljubljana}
  \country{Slovenia}
}

\author{Laurentiu Vasiliu}
\email{laurentiu.vasiliu@peracton.com}
\affiliation{%
  \institution{Peracton Ltd.}
  \country{Ireland}
}

\author{Dunja Mladenić}
\email{dunja.mladenic@ijs.si}
\orcid{0000-0003-4480-082X}
\affiliation{%
  \institution{Jo\v{z}ef Stefan Institute}
  \city{Ljubljana}
  \country{Slovenia}
}

\author{Radu Prodan}
\email{radu.prodan@uibk.ac.at}
\affiliation{%
 \institution{University of Innsbruck}
 \city{Innsbruck}
 \country{Austria}}

\author{Dumitru Roman}
\email{dumitru.roman@sintef.no}
\affiliation{%
  \institution{SINTEF AS/Bucharest University of Economic Studies}
  \country{Norway/Romania}}

%%
%% By default, the full list of authors will be used in the page
%% headers. Often, this list is too long, and will overlap
%% other information printed in the page headers. This command allows
%% the author to define a more concise list
%% of authors' names for this purpose.
% \renewcommand{\shortauthors}{Trovato et al.}

%%
%% The abstract is a short summary of the work to be presented in the
%% article.
\begin{abstract}
  Data is vital in enabling machine learning models to advance research and practical applications in finance, where accurate and robust models are essential for investment and trading decision-making. However, real-world data is limited despite its quantity, quality, and variety. The data shortage of various financial assets directly hinders the performance of machine learning models designed to trade and invest in these assets. Generative methods can mitigate this shortage. In this paper, we introduce a set of novel techniques for time series data generation (we name them Fiaingen) and assess their performance across three criteria: (a) overlap of real-world and synthetic data on a reduced dimensionality space, (b) performance on downstream machine learning tasks, and (c) runtime performance. Our experiments demonstrate that the methods achieve state-of-the-art performance across the three criteria listed above. Synthetic data generated with Fiaingen methods more closely mirrors the original time series data while keeping data generation time close to seconds - ensuring the scalability of the proposed approach. Furthermore, models trained on it achieve performance close to those trained with real-world data.
\end{abstract}

%%
%% The code below is generated by the tool at http://dl.acm.org/ccs.cfm.
%% Please copy and paste the code instead of the example below.
%%


%%
%% Keywords. The author(s) should pick words that accurately describe
%% the work being presented. Separate the keywords with commas.
%% A "teaser" image appears between the author and affiliation
%% information and the body of the document, and typically spans the
%% page.
% \begin{teaserfigure}
%   \includegraphics[width=\textwidth]{sampleteaser}
%   \caption{Seattle Mariners at Spring Training, 2010.}
%   \Description{Enjoying the baseball game from the third-base
%   seats. Ichiro Suzuki preparing to bat.}
%   \label{fig:teaser}
% \end{teaserfigure}

% \received{20 February 2007}
% \received[revised]{12 March 2009}
% \received[accepted]{5 June 2009}

%%
%% This command processes the author and affiliation and title
%% information and builds the first part of the formatted document.
\maketitle

\section{Introduction}
% How the increasing development of artificial intelligence drives the need for synthetic data to ensure we can obtain better results in data-scarce scenarios

Synthetic data has become an increasingly valuable resource for training machine learning models, offering a practical solution to the growing demand for high-quality real-world data, which is often limited, expensive, or inaccessible. This is particularly true in sensitive and data-scarce domains such as finance and healthcare, where regulatory constraints, data privacy concerns, and the complexity of data collection make synthetic data an attractive alternative.

% What is the purpose of developing machine time series generative models for finance?

Time-series data, such as stock prices, trading volumes, and economic indicators, are key to decision-making in the financial sector. Large-scale, diverse, high-frequency data resources are often required to develop robust machine learning models for financial applications. However, these data resources are rarely available in practice due to privacy concerns, market sensitivities, and licensing restrictions. As a result, there is a growing interest in generative models that can produce realistic synthetic financial time series to support research and model training.

% Challenges-> our solution

Generating realistic financial time series presents several challenges. Among them, we find challenges related to the inherent characteristics of financial time-series data (e.g., heavy-tailed distributions, interdependencies among asset returns, or nontrivial autocorrelation structures \cite{dogariu2022generation}) and, to a lesser extent, to the technical approaches used to generate them. Models such as GANs and VAEs often exhibit training instability, mode collapse, and overfitting \cite{kossale2022mode,de2025financial}. Another concern is interpretability, as many deep generative models operate as black boxes, limiting their transparency in regulated applications \cite{schwarz2024interpretable}. Moreover, while it is crucial that generative techniques capture the complex relationships and co-movements among assets \cite{masi2023correlated}, most methods assume univariate time series and independence among them. Evaluating the quality of generated sequences is also non-trivial, particularly in preserving temporal structure, diversity, and meaningful interdependencies.

To address these limitations, we propose a set of graph-based generative methods named Fiaingen \footnote{The name results from playing with words, considering \textit{Financial Data Generation (FinGen)}, the German word \textit{fangen} (\textit{to catch}), along with \text{Artificial Intelligence (AI)}, resulting in \textit{F*iai*ngen}.}. The proposed set of methods leverages the graph representations of the time series, capturing temporal dynamics and structural relationships across multiple assets, allowing for synthetic time series generation while preserving key characteristics of the original time series.

% Contributions, general to specific approach

Our approach introduces a graph-based generative framework that transforms time series into graphs. These graphs encode structural and temporal dependencies in the data and are used as the foundation for generating new, realistic synthetic sequences. This design allows us to preserve critical topological features while enabling greater control and interpretability in the generation process.

% Paper structure
The remainder of this paper is structured as follows. Section~\ref{S:MOTIVATION} presents the motivation for this work, and Section~\ref{S:DATASET} introduces the dataset with which we worked. %Section~\ref{S:METHODOLOGY} describes the methodology we followed to develop the models, 
Section \ref{S:EXPERIMENTS} describes the experiments performed. Finally, Section \ref{S:RESULTS} presents and discusses the results obtained, and Section~\ref{S:CONCLUSIONS} provides our conclusions.

\section{Motivation}\label{S:MOTIVATION}

Synthetic data, designed to replicate real-world characteristics closely, has emerged as a valuable alternative for developing and testing financial models and algorithms. Systematic Strategies LLC \cite{Syst_Strat_2023} highlights that synthetic data addresses a critical limitation of using real historical data series for modeling. Specifically, models calibrated to fit historical data often produce test results that are difficult, if not impossible, to replicate \cite{Jonathan_Kinlay_2022}. These models lack robustness against inevitable changes in dynamic statistical processes, resulting in poor out-of-sample performance. Synthetic data mitigates this issue by exposing financial models to novel scenarios, effectively stress-testing them, validating or challenging their assumptions, and revealing both strengths and weaknesses.

In a question-and-answer session conducted by Gartner \cite{Gartner_Synth_2022}, the potential of synthetic data is also highlighted, particularly to improve the accuracy of machine learning models. Real-world data is inherently incomplete and coincidental, lacking all possible conditions or event permutations. Synthetic data addresses this gap by enabling the creation of datasets encompassing edge cases and unseen conditions, expanding the scope and reliability of model testing. When implemented effectively, synthetic data offers data and analytics leaders the opportunity to develop more efficient models, potentially elevating their organisation's applications to new levels \cite{AIBusiness_Synth_2022}. Furthermore, Gartner projects that by 2030, synthetic data will surpass real-world data \cite{Gartner_Synth_2022} in its use in a wide range of machine learning models, allowing organizations to unlock the full potential of artificial intelligence technologies.

\section{Related Work}\label{S:RELATED WORK}

\subsection{Visibility Graphs and Structural Representation of Time Series}

Visibility Graphs (VGs) constitute a class of transformations that map time series into complex networks by exploiting geometric visibility criteria between observations. Two dominant variants are Natural Visibility Graphs (NVG) and Horizontal Visibility Graphs (HVG). In NVGs, nodes corresponding to time-indexed observations are connected if a straight line drawn between them does not intersect any intermediate data point, while HVGs impose a stricter horizontal visibility constraint. These mappings preserve temporal ordering and encode relative size relationships, enabling the use of graph-theoretic tools for time-series analysis.

Since their introduction, NVGs and HVGs have been extensively studied as representations of stochastic and deterministic processes \cite{Lacasa2008VisibilityGraph, Luque2009HVG}. Early analytical results demonstrated that HVGs associated with independent and identically distributed random series exhibit exponential degree distributions, while deviations from this baseline indicate temporal correlations or nonlinear structure \cite{Luque2009HVG}. Subsequent work extended visibility graph analysis to non-stationary processes, irreversible dynamics, and directed formulations capturing time asymmetry \cite{LacasaFlanagan2015}. These findings established that visibility graphs encode information about correlation structure, volatility clustering, and causal asymmetries—properties that are central to financial time series.

Despite this solid analytical foundation, the main use of visibility graphs remains descriptive. Most studies employ NVG or HVG mappings to analyse or classify existing data rather than to generate new data that preserves the encoded structural properties \cite{Mutua2015VisibilityGraph}.

\subsection{Analytical Representation and Structural Constraints}

A significant body of work over the past decade has focused on the analytical representation of visibility graph properties. Degree distributions, motif frequencies, entropy measures, and clustering coefficients have been derived or empirically characterized for broad classes of stochastic processes \cite{Luque2009HVG, Mutua2015VisibilityGraph}. Recent advances include copula-based formulations that explicitly link dependence structures in \cite{LeeJo2025CopulaHVG}. Such results provide closed-form or semi-analytical relationships between correlation strength and graph topology.

Other studies have introduced asymmetry and irreversibility measures derived from directed visibility graphs, showing strong correspondence with volatility clustering and temporal heterogeneity in financial time-series \cite{Sikorski2025Asymmetry}. These analytical developments are particularly relevant in the context of synthetic data generation, as they provide explicit structural constraints that synthetic data should satisfy in order to be statistically and dynamically consistent with real-world processes.

However, existing works mostly treat these properties as diagnostic metrics rather than as generation objectives. Structural statistics derived from visibility graphs are typically computed after data generation—often from classical stochastic models or deep generative architectures—rather than being enforced during the generative process itself.

\subsection{Multivariate, Multilayer and Multigraph Visibility Extensions}

To address the inherently multivariate nature of real-world systems, particularly financial markets, several extensions of visibility graphs have been proposed. Multilayer Horizontal Visibility Graphs (MHVGs) introduce multiple layers corresponding to individual time series, with cross-layer visibility rules encoding inter-series dependencies \cite{FreitasSilva2025MultilayerHVG}. These constructions enable the simultaneous representation of intra-series temporal structure and inter-series correlations.

Vector Visibility Graphs and related multivariate formulations further generalise the concept by treating each time step as a vector-valued observation and defining visibility criteria in a higher-dimensional space \cite{Alikasifoglu2024VISPool}. Such approaches have been explored primarily in the context of multivariate visibility graph representations, not as a generative method.

Although these multigraph and multilayer visibility frameworks significantly enhance expressive power, their application remains largely confined to analysis and supervised learning tasks. Synthetic data generation, when considered, is typically performed using external models, with visibility graphs serving only as evaluation or feature-extraction tools.

\subsection{Graph-Based and Synthetic Financial Data Generation}

There is substantial literature on synthetic financial data generation using stochastic processes, agent-based models, and deep generative architectures such as GANs, VAEs, and diffusion models. These approaches aim to reproduce marginal distributions, autocorrelation structures, and stylised facts such as fat tails and volatility clustering.

Prior work in synthetic financial data generation has explored a broad and growing literature on deep generative models. Recent systematic surveys demonstrate that Generative Adversarial Networks (GANs) and Variational Autoencoders (VAEs) dominate efforts to synthesise market time series and tabular financial data, with dozens of studies published since 2018 specifically addressing modelling fidelity and statistical realism \cite{Meldrum2025NewMoney}. Comparative reviews highlight the use of conditional GAN variants, diffusion models, and hybrid approaches to replicate stylised facts such as fat tails and volatility clustering in financial return sequences \cite{Ericson2024DeepGenTS, Takahashi2024DiffusionFinance}. Peer‑reviewed applications of GANs demonstrate synthetic scenario generation that preserves key statistical properties of multivariate financial markets, while recent work on VAE and GAN hybrids evaluates statistical similarity and quality metrics in financial contexts \cite{Ramzan2024GANFinance, Wu2023VAEFinance, Jinkou2023GANTimeSeries}.

However, purely data-driven generative models are prone to overfitting, implicit memorisation of historical trajectories, and limited interpretability. Moreover, they rarely provide explicit guarantees that higher-order temporal or cross-asset structures are preserved. This limitation is particularly problematic in regulatory, stress-testing, and AI-training contexts, where contamination from historical data and lack of structural control can invalidate downstream analyses.

Beyond domain-specific GAN and VAE architectures, recent work has explored diffusion-based conditional heteroskedastic models such as DFCH \cite{liu2024dfch}, which aim to reconstruct cleaner financial signals with realistic volatility dynamics and report improved downstream return-classification and trading performance compared to GAN-based generators and classical GARCH models. Generic diffusion architectures for long or high-dimensional time series have also been proposed, for example, TransFusion \cite{kim2024transfusion} combines transformers with diffusion to generate long sequences up to hundreds of time steps and introduces evaluation metrics tailored to both distributional fidelity and predictive characteristics, consistently outperforming GAN-based models across multiple benchmarks. Latent diffusion approaches such as TimeLDM \cite{zhang2024timeldm} first encode time series into a structured latent space using a variational autoencoder and then run the diffusion process in that latent space, achieving strong discriminative and contextual scores while improving robustness to sequence length variation.

\subsection{Limitation of Existing Literature}
Across the last decade of research, several consistent limitations emerge:
\begin{itemize}
    \item Subsequent usage of visibility graphs, with NVG and HVG representations primarily applied after data generation, rather than being integrated into the generative mechanism itself\cite{Mutua2015VisibilityGraph}.
    \item Limited scalability, despite recent progress in linear-time and online visibility graph construction algorithms\cite{HuangDeng2023LinearVG}.
    \item Weak integration with financial constraints, including regulatory compliance, historical contamination risks, and the need to preserve higher-order dependence structures.
    \item Absence of multiverse exploration, with most approaches generating a single synthetic distribution rather than a controlled family of structurally valid alternatives.
\end{itemize}

Recent advances in diffusion-based models partially address some of these gaps but also illustrate their persistence. Within finance, fully controllable diffusion models such as CoFinDiff introduce conditional architectures that incorporate conditions extracted from price data via cross-attention, enabling synthetic financial time-series generation under user-specified trajectories or regimes and thus offering explicit control over temporal and structural characteristics\cite{cofindiff2025}. In parallel, diffusion-based heteroskedastic models such as DFCH fuse GARCH-like volatility dynamics with diffusion processes to better replicate long-horizon, high-frequency financial series, reporting lower distance metrics and improved autocorrelation structure compared to QuantGAN and classical GARCH baselines\cite{liu2024dfch}. Even beyond pure generation, diffusion models are being explored as Monte Carlo replacements for simulating market dynamics more efficiently, where synthetic paths produced by diffusion pass stringent distributional tests and align with portfolio-level risk characteristics\cite{beyondmc2025}. Finally, alternative paradigms such as Quantum Wasserstein GANs with gradient penalty (QWGAN-GP) integrate quantum generators with classical discriminators to reproduce the statistical distribution of major indices (e.g., S\&P 500), showing that models trained on mixed real--synthetic data can improve forecasting of both typical conditions and extreme events, but still lack explicit mechanisms to enforce graph-based structural constraints of the type considered in this paper \cite{qwgan2024quantumsp500}.

\section{Dataset}\label{S:DATASET}

The historic financial dataset comprises daily time-series data of stock market closing prices collected over the past ten years, starting in February 2015. The data was retrieved from Yahoo Finance \cite{yahoo_finance}, which provides open access to historical market data for a wide range of financial instruments. The dataset comprises a diverse range of securities from multiple sectors, enabling us to evaluate the robustness and generalizability of synthetic data generation methods across various market conditions. It must be noticed that the data set does not cover the entire 10-year period for all securities. Some securities have shorter time-series records. Due to the large number of securities, it was most convenient to focus on just one sector and the Financial sector was considered for this work.

\section{Experiments}\label{S:EXPERIMENTS}

In this chapter, we present the experiments conducted to evaluate the performance of our proposed data generation method in comparison with several state-of-the-art synthetic time-series generation approaches. The primary objective is to assess the quality and usability of the synthetic data generated by our model against existing techniques such as TimeGAN, conditional GAN (cGAN), Structural Time Series Generator (STS), and DiffusionTS. In addition, we consider the Values Random Permutation (VRP), where the values in each time series window are randomly shuffled, and use it for ablation studies against our proposed methods. 

All experiments for synthetic time series generation were performed on a remote server running a 64-bit Linux system (Ubuntu, x86\_64 architecture) equipped with dual Intel(R) Xeon(R) E5-2630 CPUs (2 sockets, 12 cores total, 24 threads) clocked at 2.30GHz and 24 logical processors. The experiments were conducted in a controlled environment to ensure reproducibility, with all dependencies version-pinned. Due to compatibility constraints with TensorFlow 1.15.0, TimeGAN was executed in a dedicated Python 3.6 environment, isolated from the main Kedro pipeline (Python 3.10), which uses TensorFlow 2.15.1-compatible packages. Inter-environment communication was handled via subprocess execution and file-based I/O to ensure full reproducibility. The code used to execute the experiments was made available in the following repository: https://github.com/the-repository-url-will-be-updated-upon-acceptance.

For the cGAN and STS synthetic time series generation, we used the tsgm library \cite{nikitin2023tsgm}. Furthermore, we incorporated DiffusionTS, a state-of-the-art diffusion-based method \cite{yuan2024diffusionts} and the TimeGAN model \cite{NEURIPS2019_c9efe5f2} as referenced in the corresponding papers.

We conducted a comparative analysis by generating synthetic time-series data with each of these methods and comparing them against our proposed method. To evaluate the ability of each approach to replicate the underlying temporal dynamics and statistical characteristics of real historical financial time-series data, we employed three evaluation criteria: (a) comparison of t-SNE visualizations to inspect the similarity of the synthetic and real datasets in the latent space, (b) outcomes on classification tasks, and (c) runtime performance assessment. Regarding classification, we trained machine learning models on three variants of the datasets: (i) real historical data, (ii) synthetic data produced by each generative method, and (iii) a mixed dataset combining real and synthetic data. This evaluation aimed to understand how well the synthetic data preserves information relevant to downstream tasks.

Given that some of the generative methods were highly time-consuming, we generated a reduced number of them and, subsequently, constrained the analyses to the resulting dataset sizes (e.g., 255 tickers in general, and 160 tickers when considering the synthetic data generated with Diffusion models).

These experiments provide comprehensive insights into the fidelity, diversity, and real-world applicability of the generated data. They also help to highlight the comparative advantages and limitations of our method in relation to state-of-the-art methods, such as TimeGAN \cite{NEURIPS2019_c9efe5f2}, cGAN \cite{mirza2014conditional,smith2020conditional}, STS \cite{nikitin2023tsgm}, and DiffusionTS \cite{yuan2024diffusionts}.

\subsection{Synthetic data generation}

\subsubsection{TimeGAN}
TimeGAN is a generative time-series model, trained adversarially and jointly via a learned embedding space with both supervised and unsupervised losses. TimeGAN consists of four network components: an embedding function, a recovery function, a sequence
generator, and a sequence discriminator. The key insight is that the autoencoding components (first two) are trained jointly with the adversarial components (latter two), such that TimeGAN simultaneously
learns to encode features, generate representations, and iterate across time. The embedding network
provides the latent space, the adversarial network operates within this space, and the latent dynamics
of both real and synthetic data are synchronized through a supervised loss \cite{NEURIPS2019_c9efe5f2}. 

For the experiment, we created smaller time series by 'slicing' the original time series from the dataset into segments of a particular size. The training process was applied independently to each time series in the dataset. Before feeding the time series to the model, we did some data pre-processing, such as filtering out windows with missing values, scaling data with a Min-Max function, and randomly shuffling it. The TimeGAN model was configured considering a hidden dimension of 24 units, 3 recurrent layers, and a batch size of 128 and trained for 1,000 epochs. Generated sequences were inverse-transformed to the original scale to enable a fair comparison between real and synthetic sequences.

Lastly, to ensure consistency in evaluation across methods and avoid bias due to imbalanced sequence counts, we randomly downsampled the generated synthetic sequences per time series to a fixed number, to ensure a uniform synthetic dataset size across models. In addition, to consider the best-case scenario, were the generated synthetic time series would best resemble the original ones, we conducted a second selection considering those most similar to the original time series using Dynamic Time Warping (DTW). This helped ensure that the selected synthetic sequences preserved the structural characteristics of the real data. We note these two data selection strategies as DS (random sample selection) and SimDS (similarity-based sampling).

\subsubsection{cGAN}

Conditional GAN (cGAN) extends the traditional Generative Adversarial Network framework by conditioning both the generator and discriminator on additional information, such as class labels or input features. In the context of time-series generation, cGAN learns to produce sequences by incorporating conditional signals—typically timestamps or cluster labels—allowing for controlled generation of time-dependent patterns. The generator is trained to produce synthetic sequences that align with the given condition, while the discriminator aims to distinguish between real and synthetic sequences, also considering the condition. This conditioning mechanism helps guide the generation process, improving the relevance and fidelity of the generated time-series data to the target distribution.

In our experiments, we divided the historical data into sequences, each of which represented a segment of the time series. We trained the cGAN model under different conditions and for varying numbers of epochs (500, 600, and 700). The conditions tested included:

\begin{itemize}
    \item Whether the slope of the data was positive (labeled as 1) or negative (labeled as 0),
    \item Whether the price increased or decreased from the 19\textsuperscript{th} to the 20\textsuperscript{th} time step,
    \item The volatility of each time window,
    \item The Relative Strength Index (RSI) of the entire time window, and
    \item 
The overall trend was represented as the slope of the whole window.
\end{itemize}

We observed that the model was unstable, with results fluctuating significantly—even for the same condition and number of epochs—sometimes performing very well and at other times performing poorly. After further experimentation, we decided to use the RSI of the entire time window as the sole condition.

The model architecture consisted of a generator and discriminator from the \texttt{tsgm} model zoo (\texttt{cgan\_base\_c4\_l1}), configured for one-dimensional time-series sequences.

We used the Adam optimizer with a learning rate of $0.0002$ and $\beta_1 = 0.5$ for both generator and discriminator. The model was trained for 700 epochs using batches of size 32. During training, we monitored generation using a custom callback class that periodically sampled synthetic sequences.

\subsubsection{STS}
Structural Time Series (STS) models are designed to capture the underlying components of a time series—such as trend, seasonality, and noise—by explicitly modeling them as part of the generative process. This decomposition enables the generation of more interpretable and realistic synthetic sequences, as each component contributes to the final structure of the time series.

We applied this model to sequences of length 20, using each sequence as input to forecast the subsequent 20 time steps. To select the optimal model configuration, we tested different settings on a representative time series: varying the number of variational steps (\texttt{num\_variational\_steps} = 100, 150, 175, 200) and comparing two trend models—autoregressive (AR(1)) and local linear trend. Based on the quality of the generated sequences visualized using t-SNE plots, we selected the AR(1) trend with 150 variational steps as the most effective setup. To improve efficiency, we trained the model independently on each input window using parallel processing and stored the generated synthetic sequences for each time series column in separate files. Execution times were logged per column to monitor scalability.

\subsubsection{Diffusion TS}
Diffusion-TS is a diffusion-based generative framework designed to model and synthesize general time series data, supporting both conditional and unconditional generation. The architecture consists of two primary components: a sequence encoder and an interpretable decoder. The encoder captures temporal dependencies using a stack of attention and feed-forward layers, configured with multi-head self-attention and residual connections. The decoder explicitly decomposes the reconstructed time series into trend and seasonal components. The trend is modeled using a combination of polynomial regression and the averaged block outputs from the decoder layers. The model employs Fourier-based trigonometric functions for the seasonal component, capturing periodic patterns efficiently.

We used the implementation that uses a cosine beta schedule across 500 diffusion timesteps, both during training and sampling. It is trained using L1 loss on sequences of length 20 with a hidden feature size of 64. The training is performed for 2,500 epochs, using a ReduceLROnPlateau scheduler and exponential moving average (EMA) for stabilization. The dataset is normalized to the range of -1 to 1, and the model was trained on a batch size of 64.

Finally, to ensure consistent evaluation and eliminate bias from uneven sequence counts, we used the \texttt{downsample\_sequences} function to reduce the number of synthetic sequences generated by DiffusionTS to a fixed amount per column. When more sequences were available than needed for a single time series window, we followed the same sample selection procedure as described for TimeGAN, considering a random sampling selection (DS) or a set of samples closest to the original ones, considering the DTW distance (SimDS).

\subsubsection{Fiaingen (our method)}

We transformed time series segments into visibility graphs using several strategies to explore the structural properties of the time series and support downstream graph-based learning methods. The transformation was applied to windows of length 20 and, for comparison, also to longer segments of length 60.

We experimented with the following graph construction strategies:

\begin{itemize}
    \item Natural Visibility Graphs (NVG) \cite{lacasa2008time,silva2021time}: Captures visibility between data points based on the convexity criterion of the time series curve.

    \item Horizontal Visibility Graphs (HVG) \cite{luque2009horizontal,silva2021time}: A simplified variant of NVG, where visibility is determined using a horizontal line-of-sight rule.

    \item Multigraphs with Natural Visibility Strategy: Combines NVGs created from time series segments corresponding to the same time window into a single composite graph. For each time window, a multigraph is built where each ticker’s sequence contributes as an individual time series NVG and linked using a time co-occurrence linking method and further refined by linking nodes with similar values and combining identical nodes.
\end{itemize}

We generate time series from the graphs described above by walking through the graph and appending values to a resulting sequence. The process ensures that the structural relationships represented within the graph are preserved in the reconstructed time series. There are two key moments required for the timeseries generation: selecting the next node in the graph and selecting the associated value, which we describe below: selection of the next node in the graph and selection of the next value in the selected node.  

\textit{Selection of the next node in the graph} Several strategies are employed to select the next node while generating time series from graphs, each offering a distinct approach to navigating the graph's structure. Random selection is the simplest strategy, where the next node is chosen randomly from all appropriate nodes. For multivariate graphs, neighborhood-based strategies consider transitions to neighbors either within a single graph or across multiple univariate graphs. Some approaches, like random neighbor selection, ensure systematic exploration, while others, such as random neighbor selection with graph switching, alternate between graphs to capture cross-graph dynamics. Restart-based strategies will choose the next node randomly, but they will always consider a certain percentage (in our case, 15\%) chance that a choice will be made to jump back to the first node of the graph. Finally, degree-weighted strategies select the next node based on the degree/weight of the connection between the two, prioritizing highly connected nodes for transitions. These strategies provide flexibility and customization in time series reconstruction, depending on the graph's structure and desired application. In our case, when synthesizing new data for the experiments, we considered random walks with restarts.

\textit{Selection of the next value in the selected node} 
Some strategies for selecting the next value in a node handle situations where nodes contain multiple values rather than a single one. Random selection methods choose a value from the available ones without preference, while round-robin strategies sequentially cycle through values in the order they appear, starting over when the end is reached.

\begin{figure}[ht]
\centering
\includegraphics[width=\linewidth]{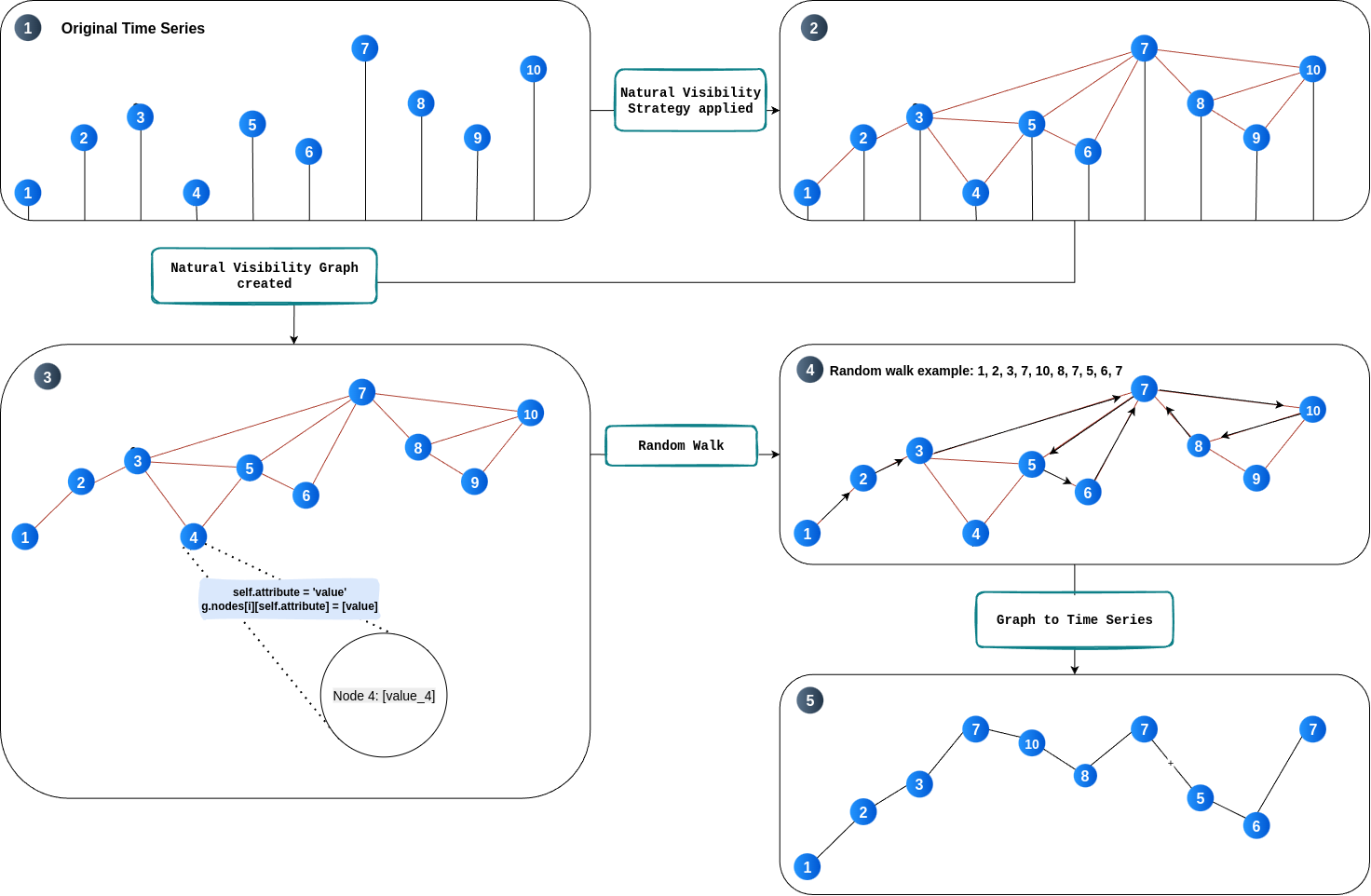}
\caption{Diagram detailing the generative process considering NVG.}
\label{fig:ts2g2_appendix_diagram}
\end{figure}

We illustrate the approach for NVG in Fig. \ref{fig:ts2g2_appendix_diagram}.

\subsubsection{Ablation study: Values Random Permutation (VRP)}

To understand how much of the performance of the method we propose can be attributed to the time series topological properties encoded in a graph, we implemented a random permutation generator. This baseline generates a synthetic time series by applying a random shuffle to each window of size 20 or 60 of the original data. Unlike generative models that aim to learn temporal dependencies and dynamics, this method randomly changes the order of values in each time series window. While the values themselves stay the same, any patterns or trends over time are lost.

\subsection{Performance assessment on downstream tasks: classification}

We compared the performance of a model trained on synthetic data versus one trained on historical data, as well as a mixed dataset containing both synthetic and historical data. The closing prices used in the classification are generated by the models from historical data and then sliced into smaller segments. Classification is performed by using the data from the first 19 closing prices and computing features over them (e.g., linear trend, polynomial trend, average change, RSI, number of peaks, mean, variance, among others), while the target label is determined by the direction of the 20th closing price: the label is 1 if the 20th price goes up, and 0 if it goes down. Additional features are also included to further enrich the input data.

For the classification task, we use the CatBoost classifier \cite{catboost}, which has been shown to perform well with categorical data. We train separate models on three different datasets: one on the historical data, one on the synthetic data, and one on a mixed dataset that combines both synthetic and historical data. To evaluate the performance of the classifier, we trained the model on the training set and then evaluated it on the validation and test sets. The performance was measured using the ROC AUC metric. The test set always contains real (historical) data, ensuring that the model is evaluated on its ability to generalize to unseen, actual data. The purpose of the classification task is to evaluate how the classifier performs on different synthetic data, allowing us to compare the quality of data generated by our method with the benchmark method.

As mentioned in the previous sections, for both TimeGAN and DiffusionTS, we performed classification using synthetic data that was downsampled with two different techniques: random sampling and similarity-based selection using Dynamic Time Warping (DTW). This allowed us to assess how the choice of downsampling method affects the downstream classification performance.

\section{Results and Discussions}\label{S:RESULTS}

In this chapter, we present the results of our experiments, focusing on the performance of different generative methods for synthetic time-series data generation. Specifically, we evaluate the results using t-SNE plots to visualize the distribution of generated data and the classification performance on synthetic data generated by both methods.

The presented results aim to evaluate the effectiveness of the proposed generative methods in producing realistic synthetic data for financial applications. For clarity, the best-performing values in each table are highlighted in bold.

\begin{figure*}[htbp]
    \centering

    % Row 1
    \begin{subfigure}{0.23\textwidth}
        \includegraphics[width=\linewidth]{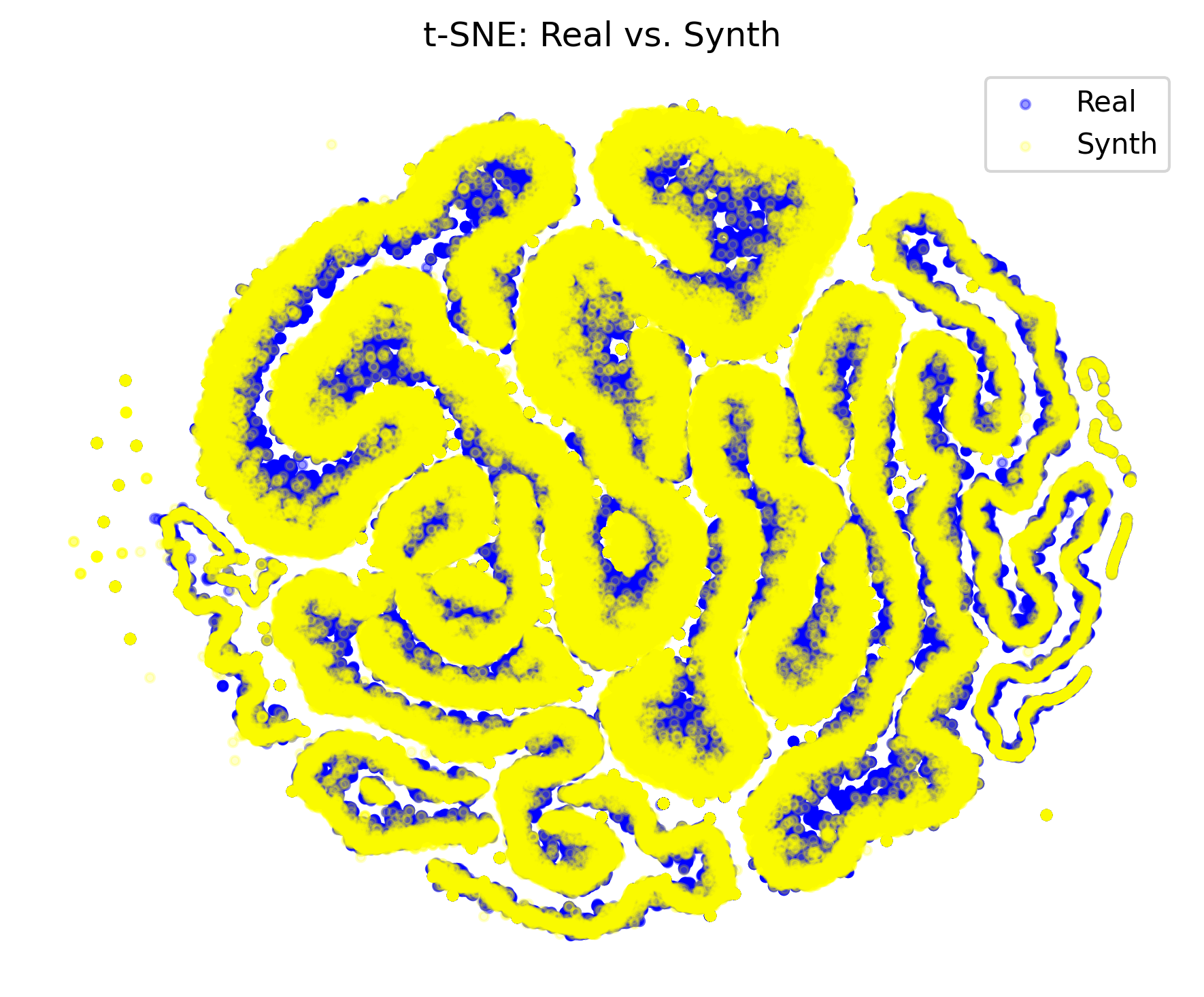}
        \caption{NVG}
        \label{fig:ts2g2_natural_20}
    \end{subfigure}
    \hfill
    \begin{subfigure}{0.23\textwidth}
        \includegraphics[width=\linewidth]{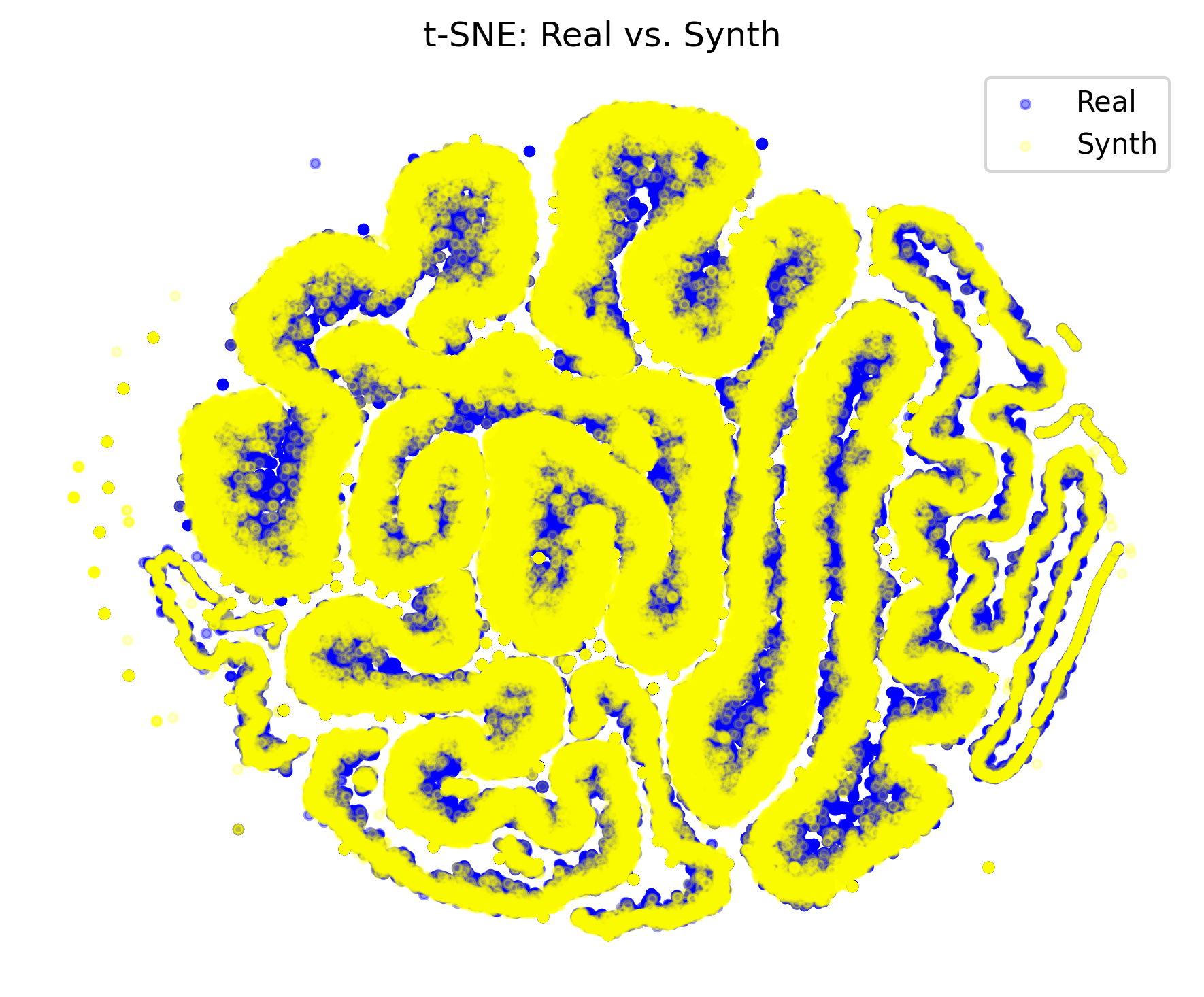}
        \caption{HVG}
        \label{fig:ts2g2_horizontal_20}
    \end{subfigure}
    \hfill
    \begin{subfigure}{0.23\textwidth}
        \includegraphics[width=\linewidth]{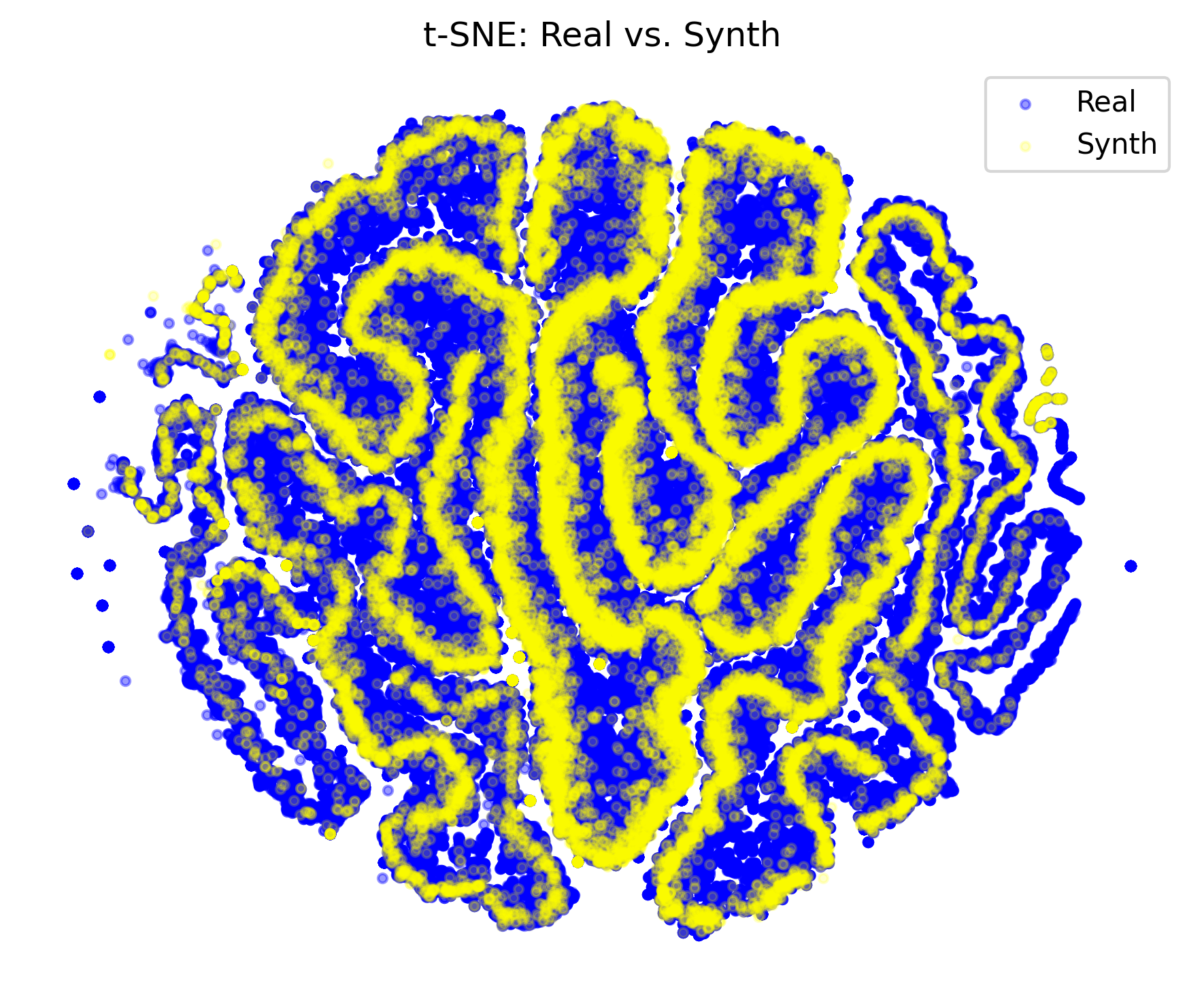}
        \caption{Multigraph}
        \label{fig:ts2g2_multi_20}
    \end{subfigure}
    \hfill
    \begin{subfigure}{0.23\textwidth}
        \includegraphics[width=\linewidth]{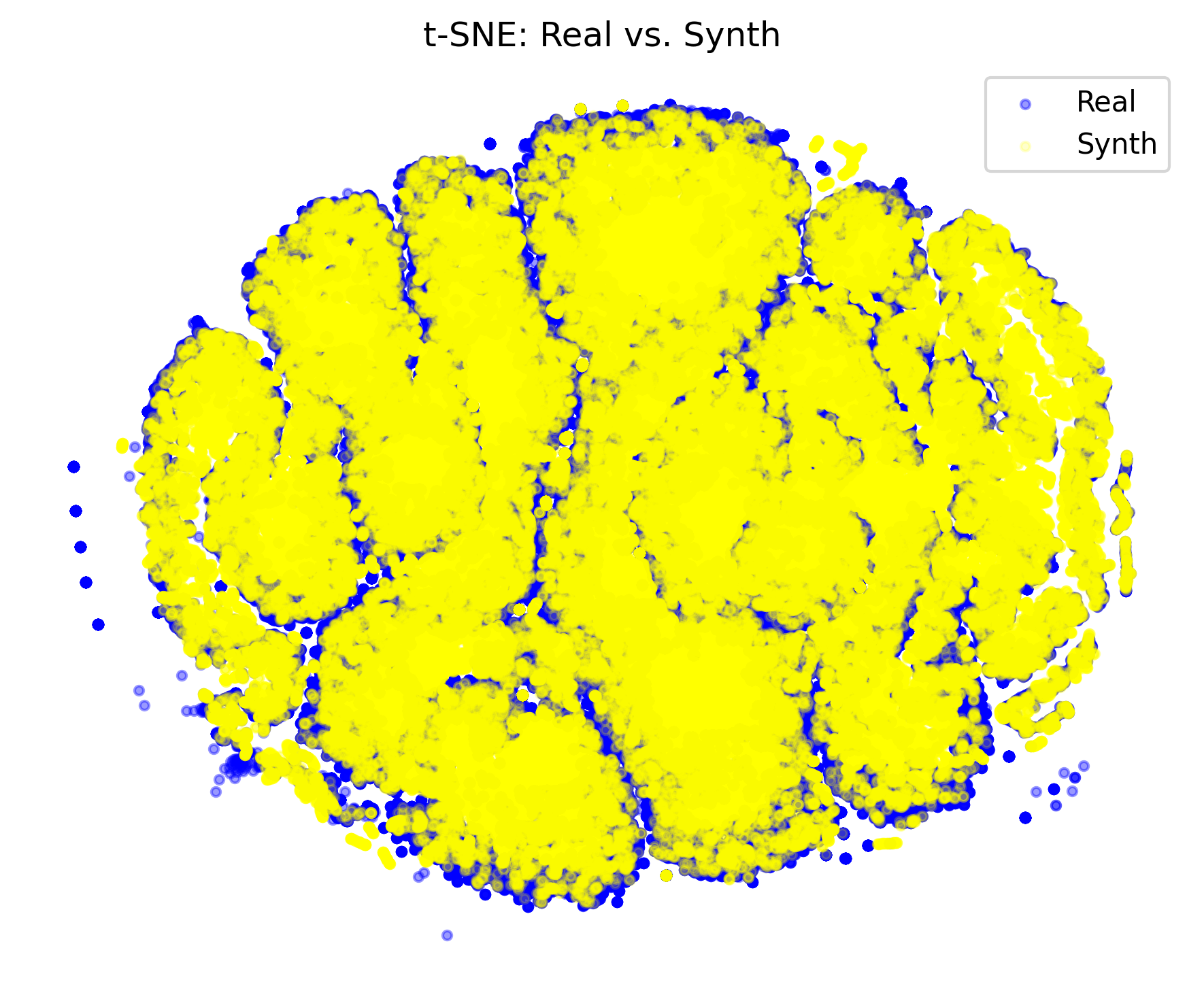}
        \caption{TimeGAN}
        \label{fig:timegan_20}
    \end{subfigure}

    \vspace{1em}

    % Row 2
    \begin{subfigure}{0.23\textwidth}
        \includegraphics[width=\linewidth]{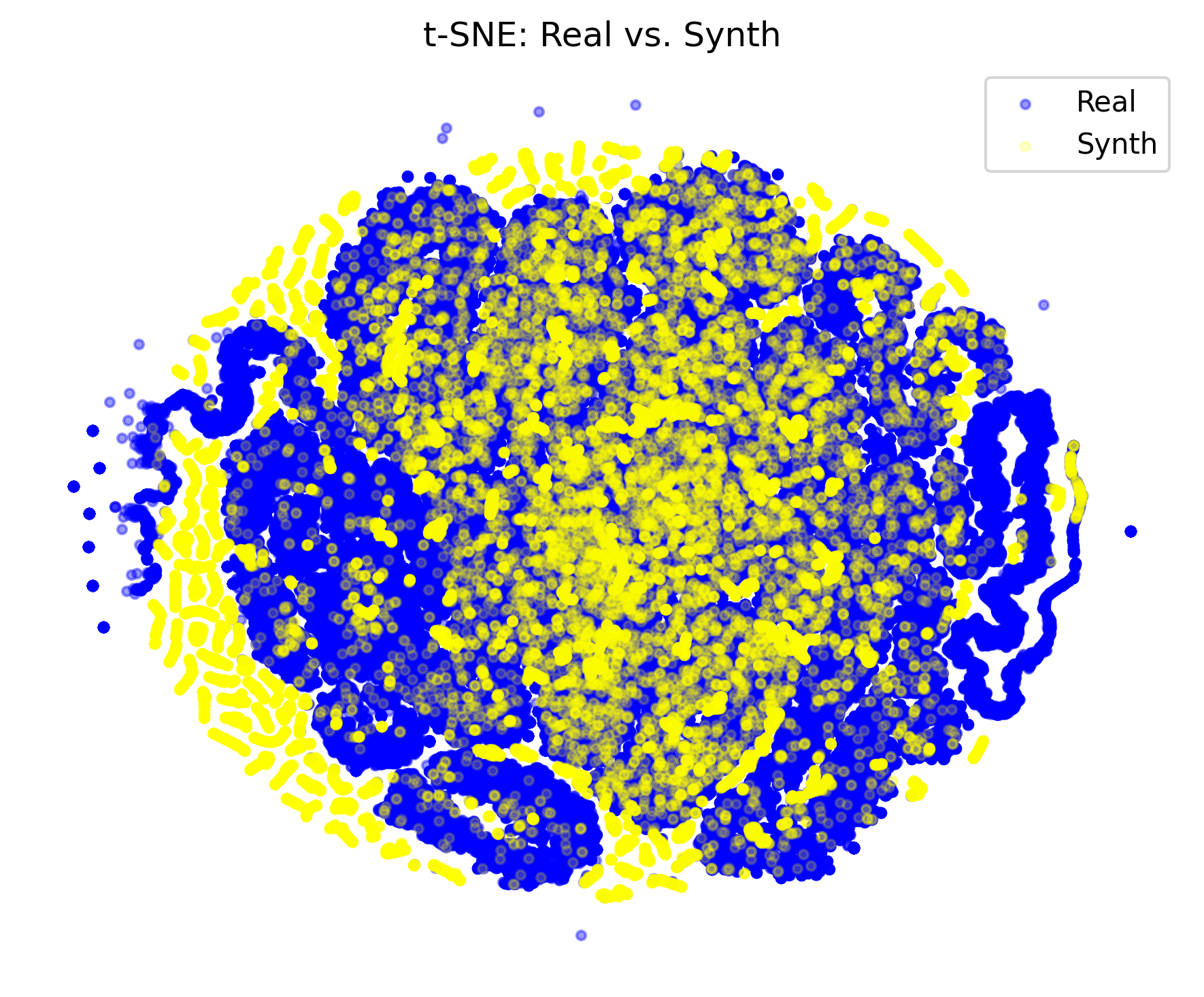}
        \caption{cGAN}
        \label{fig:cgan_20}
    \end{subfigure}
    \hfill
    \begin{subfigure}{0.23\textwidth}
        \includegraphics[width=\linewidth]{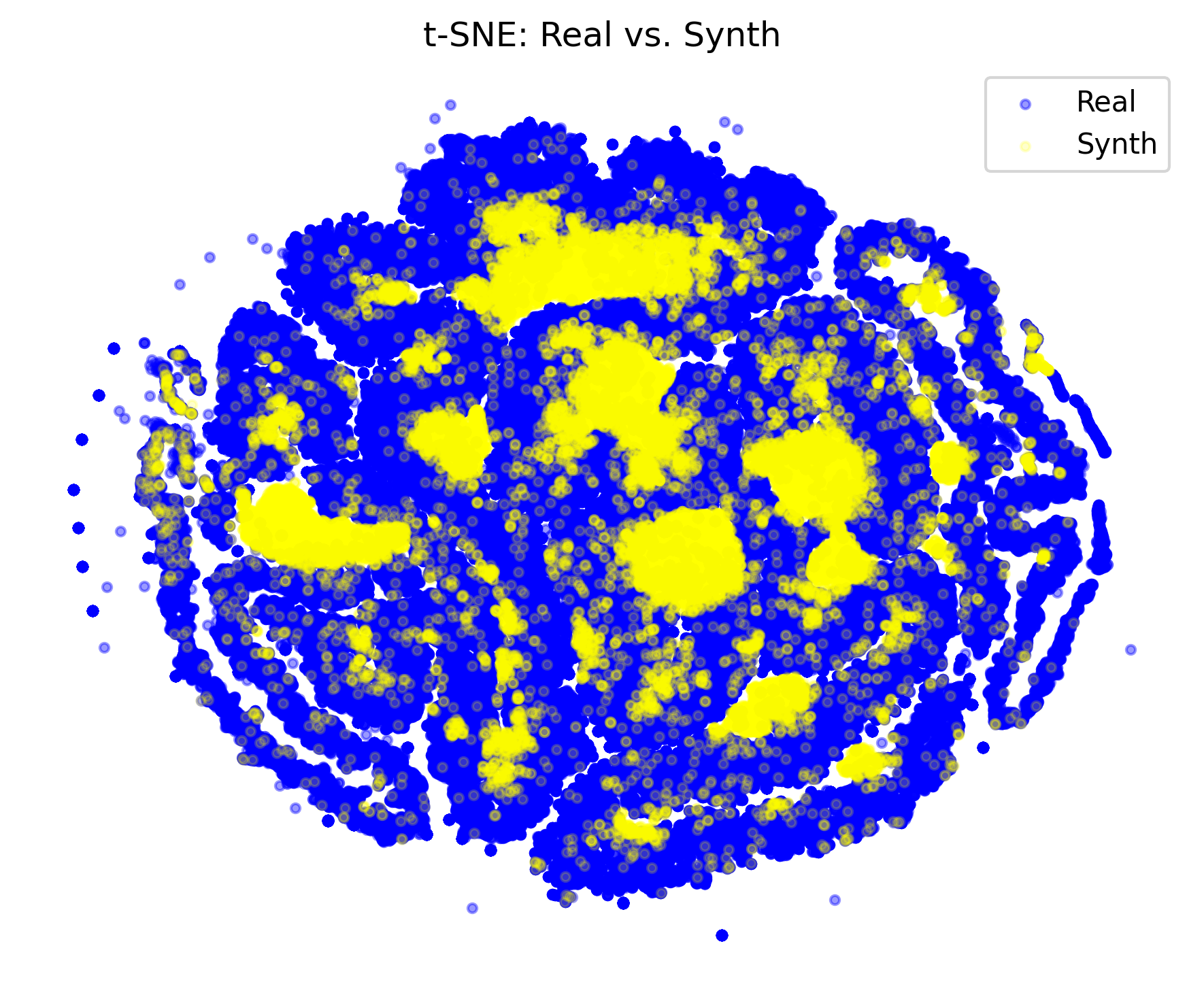}
        \caption{STS}
        \label{fig:sts_20}
    \end{subfigure}
    \hfill
    \begin{subfigure}{0.23\textwidth}
        \includegraphics[width=\linewidth]{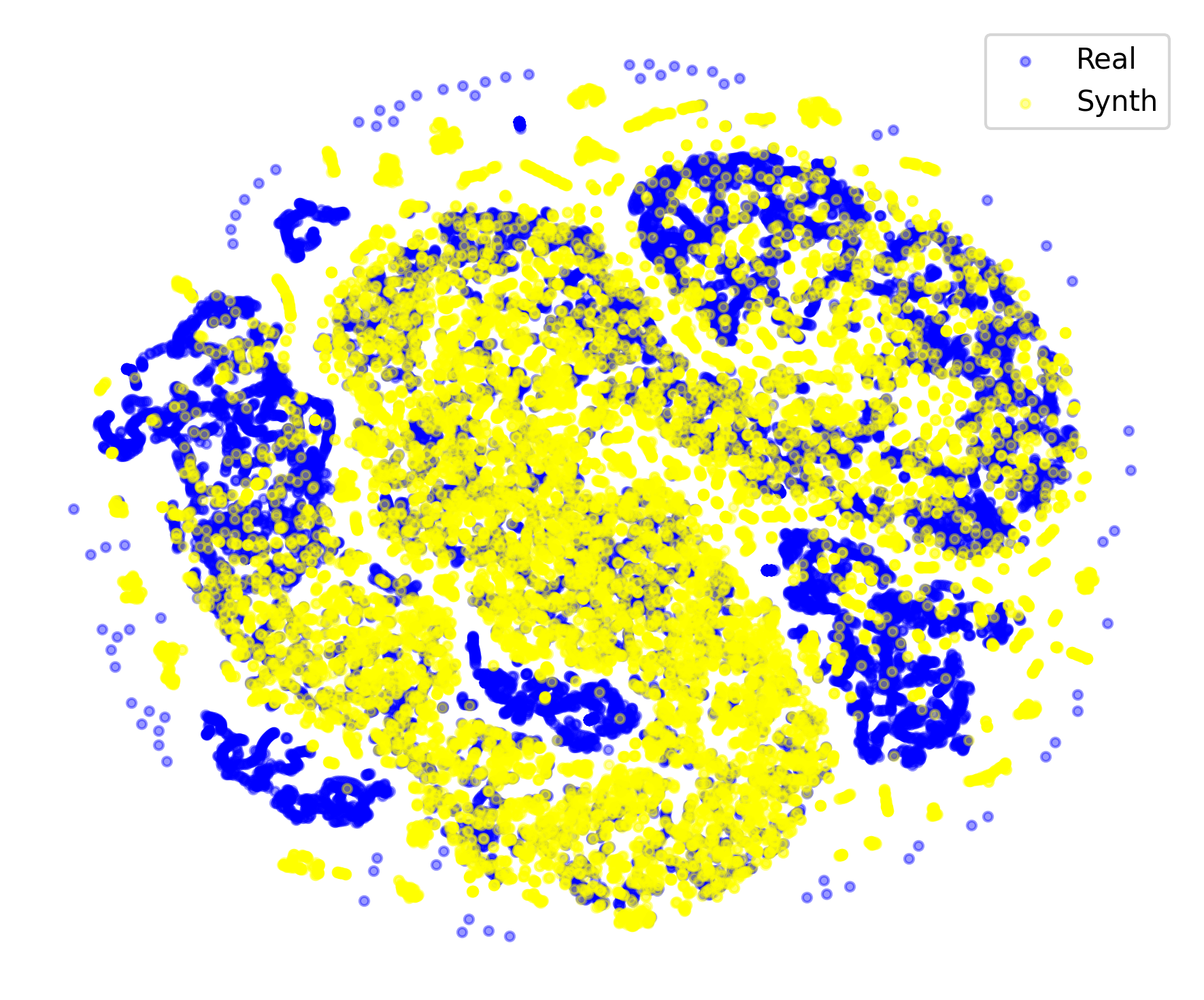}
        \caption{DiffusionTS}
        \label{fig:diffusionts_20}
    \end{subfigure}
    \hfill
    \begin{subfigure}{0.23\textwidth}
        \includegraphics[width=\linewidth]{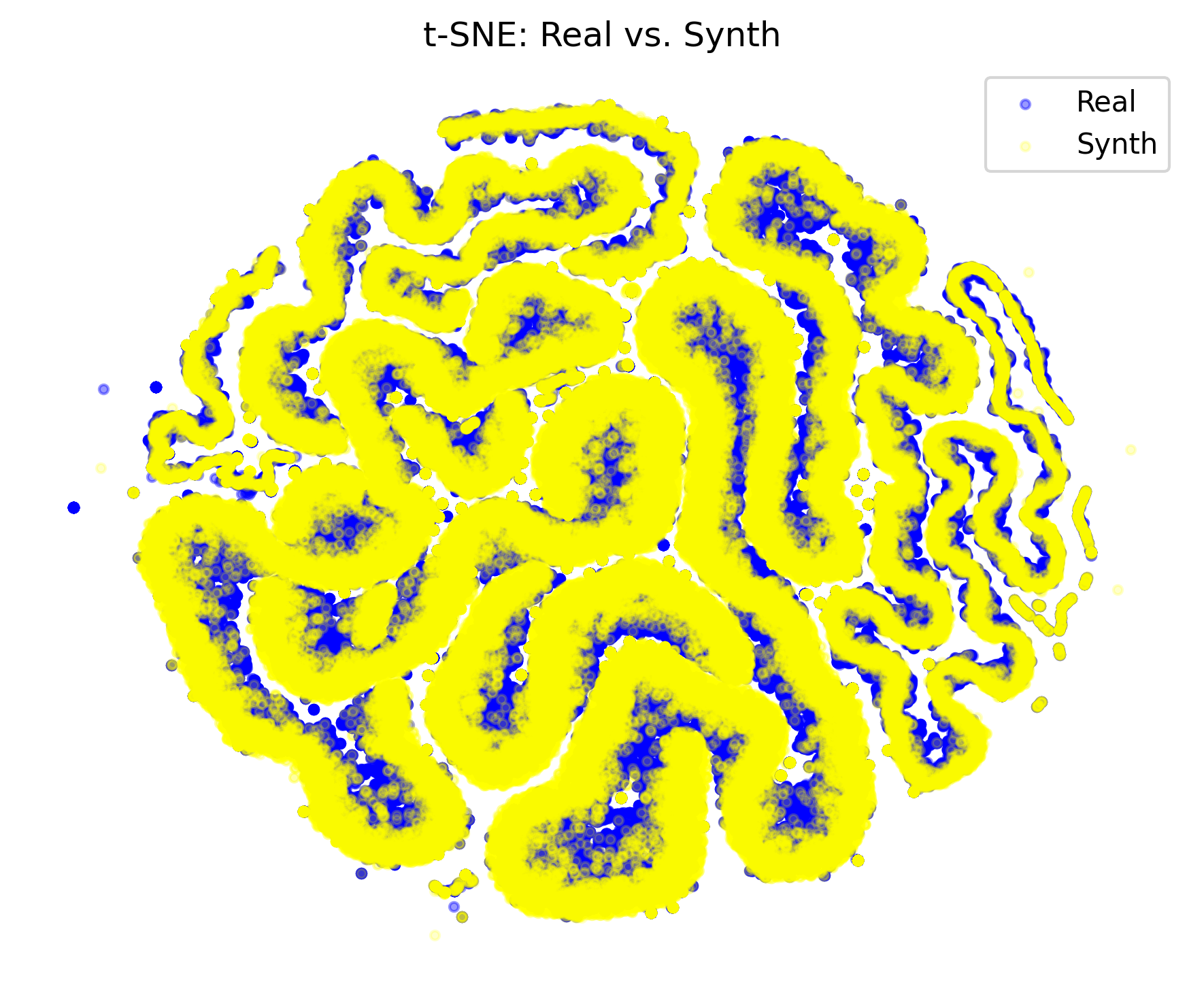}
        \caption{Values Random Permutation}
        \label{fig:random_20}
    \end{subfigure}

    \caption{t-SNE visualizations comparing real vs. synthetic time series for each generative strategy (window size = 20). Blue: real, Yellow: synthetic.}
    \label{fig:tsne_grid_win20}
\end{figure*}

\begin{figure}[htbp]
    \centering

    \begin{subfigure}{0.30\textwidth}
        \includegraphics[width=\linewidth]{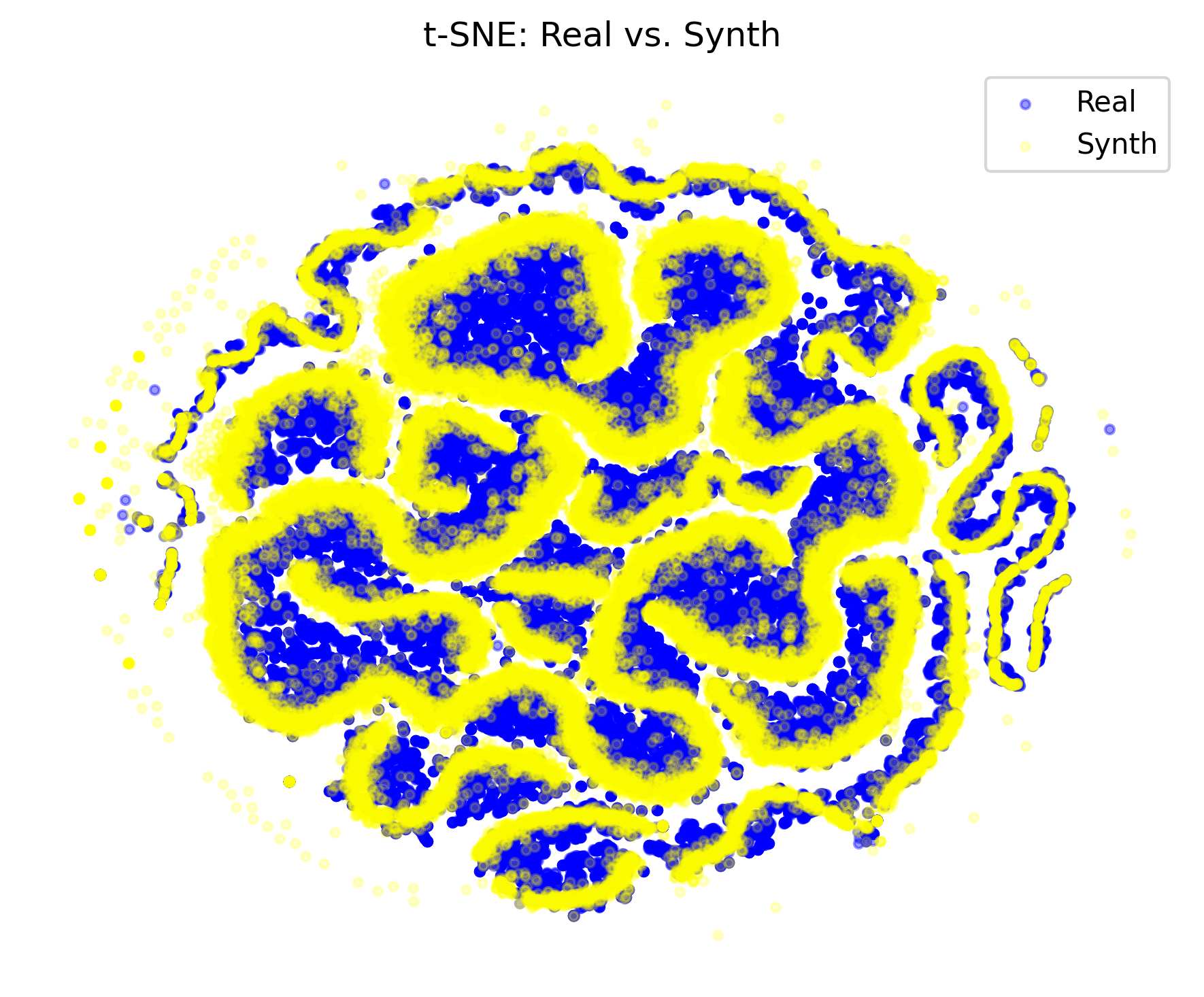}
        \caption{NVG}
        \label{fig:ts2g2_natural_60}
    \end{subfigure}
    \hfill
    \begin{subfigure}{0.30\textwidth}
        \includegraphics[width=\linewidth]{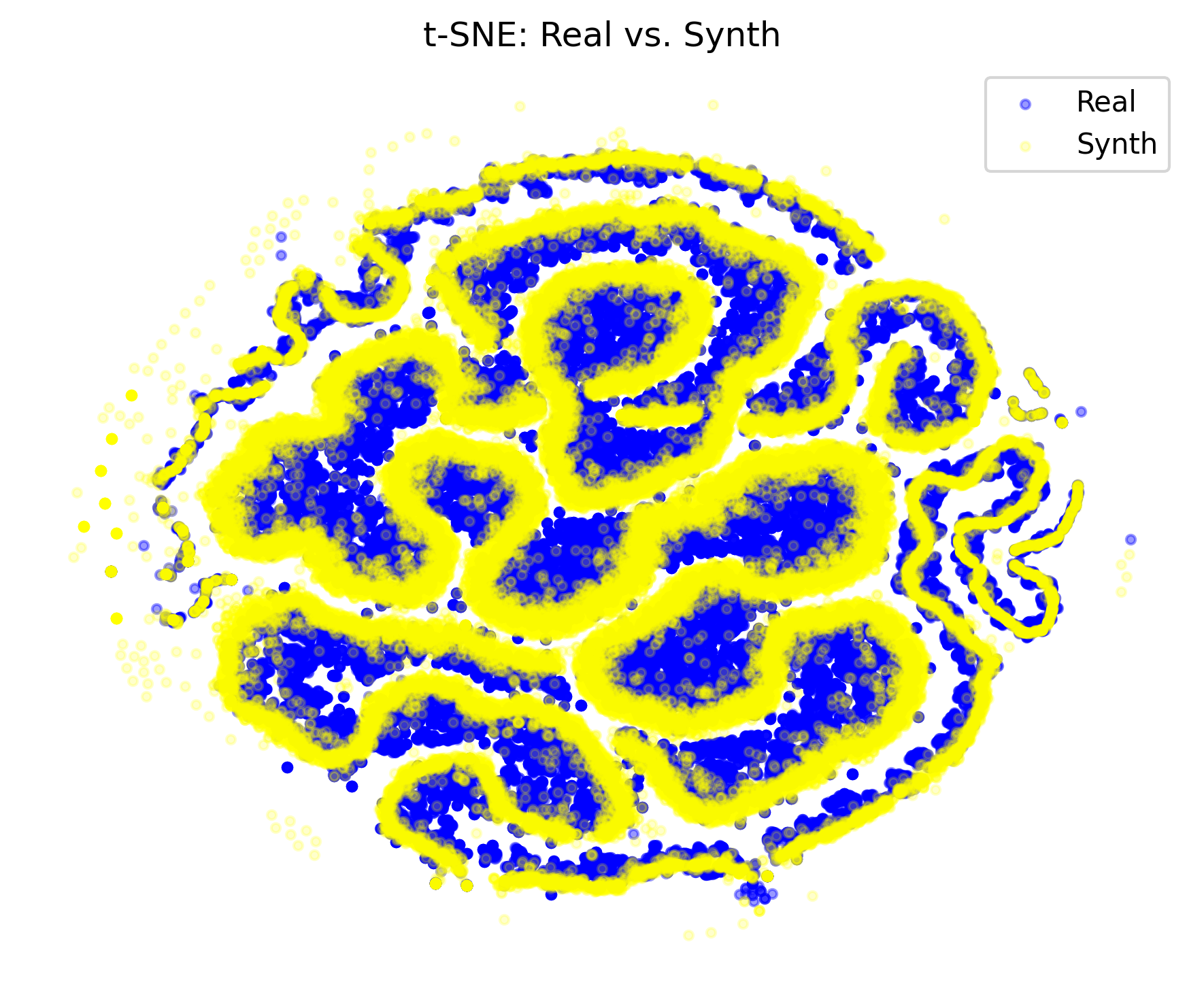}
        \caption{HVG}
        \label{fig:ts2g2_horizontal_60}
    \end{subfigure}

    \caption{t-SNE visualizations comparing real vs. synthetic time series for NVG and HVG generative strategies (window size = 60). Blue: real, Yellow: synthetic.}
    \label{fig:tsne_appendix_60}
\end{figure}

\subsection{Data Quality Assessment with t-SNE plots}

The t-SNE plots provide a visual representation of the high-dimensional synthetic data in a lower-dimensional space, allowing us to explore the separation and clustering of the generated data. As such, they are frequently used to assess the quality of synthetic data \cite{ang2023tsgbench}.

To evaluate the generated data, t-SNE plots were created for our proposed methods and all benchmark methods. In our method, the plots were generated for synthetic data produced using two time window sizes for graph creation: 20 and 60. Figures~\ref{fig:ts2g2_natural_20} and~\ref{fig:ts2g2_natural_60} show the t-SNE plots for data generated using the natural visibility graph strategy, while Figures~\ref{fig:ts2g2_horizontal_20} and~\ref{fig:ts2g2_horizontal_60} show results for the horizontal visibility graph strategy.

These plots allow us to visually inspect how closely the synthetic data matches the real data in terms of clustering and distribution. Substantial overlap or intermixing of real and synthetic points indicates that the generative model has successfully captured the underlying structure of the original time series data.

Our proposed graph-based strategies, which are the NVG (with window sizes 20 and 60), multigraph (window size 20), and horizontal visibility (window size 60),  demonstrate strong intermixing between real and synthetic data points, indicating high fidelity in preserving the underlying temporal and structural properties. 

We expected the Values Random Permutation method (Figure~\ref{fig:random_20}) to show a strong overlap between real and synthetic data in the t-SNE projection, given that the same values from the original time series are used. Nevertheless, the overlap can be deceiving, given that the t-SNE plots provide no evidence on the topological properties of the time series.

In contrast, benchmark models such as cGAN, STS, and DiffusionTS exhibit noticeable separation between real and synthetic data values, suggesting limitations in capturing the full complexity of the original data. TimeGAN performs moderately well, achieving partial overlap but still showing distinct clusters.

These qualitative results support the effectiveness of our visibility-based framework for generating realistic financial time series, while also highlighting the need for caution when interpreting similarity in visualization methods like t-SNE.

\subsection{Performance on downstream tasks: classification results}
In the classification task, we compared how different training sets affected the predicted outcomes of the classification model predicting whether the n\textsuperscript{th} price will go up or down when considering a time series segment of n values. This provides insights into the models' ability to produce realistic and usable synthetic data for further analysis and modeling. It must be noticed that the test set was formed with real-world data only and remained the same across experiments. Table~\ref{tab:combined_auc_table} presents the classification results on synthetic data generated for 255 different stocks using all methods except DiffusionTS. In contrast, Table~\ref{tab:combined_auc_table_160} shows results for 160 stocks, including DiffusionTS as well.

\begin{table*}[ht]
\centering
\footnotesize
\begin{tabular}{|c|c|c|c|c|c|c|c|c|}
\hline
\textbf{Dataset} & \textbf{NVG} & \textbf{HVG} & \textbf{NVMG} & \textbf{TimeGAN-DS} & \textbf{TimeGAN-SimDS} & \textbf{cGAN} & \textbf{STS} & \textbf{VRP (Ablation)} \\ \hline
Real             & 0.85212      & 0.85212      & 0.85212       & 0.85212             & 0.85212                & 0.85212       & 0.85212      & 0.85212         \\ \hline
Synthetic        & 0.74520      & \textbf{0.74891} & 0.74613       & 0.60665             & 0.59581                & 0.74699       & 0.63390      & 0.71443         \\ \hline
Mixed            & 0.84040      & \textbf{0.84095} & 0.83920       & 0.80136             & 0.80691                & 0.83356       & 0.80652      & 0.83661         \\ \hline
\end{tabular}
\caption{ROC AUC obtained for different training sets (Real, Synthetic, Mixed) using various synthetic data generation methods on a dataset of 255 tickers: NVG, HVG, NVMG, TimeGAN with random downsampling and TimeGAN with similarity-based downsampling (TimeGAN-DS, TimeGAN-SimDS), cGAN, STS, and a Random Permutation baseline.}
\label{tab:combined_auc_table}
\end{table*}

\begin{table*}[ht]
\centering
\footnotesize
\begin{tabular}{|c|c|c|c|c|c|c|c|c|c|c|}
\hline
\textbf{Dataset} & \textbf{NVG} & \textbf{HVG} & \textbf{NVMG} & \textbf{TimeGAN-DS} & \textbf{TimeGAN-SimDS} & \textbf{cGAN} & \textbf{STS} & \textbf{DiffusionTS-DS} & \textbf{DiffusionTS-SimDS} & \textbf{VRP (Ablation)} \\ \hline
Real             & 0.82366      & 0.82366      & 0.82366       & 0.82366             & 0.82366                & 0.82366       & 0.82366      & 0.82366                  & 0.82366                    & 0.82366         \\ \hline
Synthetic        & \textbf{0.73071}     & 0.72598      & 0.71841       & 0.59583             & 0.61375                & 0.66635       & 0.62778      & 0.61257                  & 0.62082                    & 0.722           \\ \hline
Mixed            & 0.80331      & \textbf{0.81206}      & 0.80053       & 0.75867             & 0.73463                & 0.79626       & 0.77936      & 0.73136                  & 0.75252                    & 0.79514         \\ \hline
\end{tabular}
\caption{ROC AUC results for different datasets (Real, Synthetic, Mixed) using various synthetic data generation methods on a dataset of 160 tickers: NVG, HVG, NVMG, TimeGAN with random downsampling and TimeGAN with similarity-based downsampling (TimeGAN-DS and TimeGAN-SimDS), cGAN, STS, DiffusionTS with random downsampling (DiffusionTS-DS), DiffusionTS with similarity-based downsampling (DiffusionTS-SimDS), and the VRP baseline.}
\label{tab:combined_auc_table_160}
\end{table*}

\begin{table}[ht]
\centering
\footnotesize
\begin{tabular}{|c|c|c|c|}
\hline
\textbf{Dataset} & \textbf{NVG} & \textbf{HVG} & \textbf{VRP (Ablation)} \\ \hline
Real             & 0.90501             & 0.90501             & 0.90501                \\ \hline
Synthetic        & \textbf{0.73463}    & 0.72972             & 0.70947                \\ \hline
Mixed            & \textbf{0.90189}    & 0.90090             & 0.89849                \\ \hline
\end{tabular}
\caption{ROC AUC results using input window of length 60. Graph-based methods (NVG and HVG) are compared to the Random Permutation baseline for Real, Synthetic, and Mixed datasets.}
\label{tab:graph_auc_window_60}
\end{table}

\begin{table}[ht]
\centering
\footnotesize
\begin{tabular}{|c|c|c|c|}
\hline
\textbf{Dataset} & \textbf{NVG} & \textbf{HVG} & \textbf{VRP (Ablation)} \\ \hline
Real             & 0.93629             & 0.93629             & 0.93629                \\ \hline
Synthetic        & \textbf{0.85495}    & 0.85133             & 0.82868                \\ \hline
Mixed            & \    0.92954    & \textbf{0.93023}            & 0.92619                \\ \hline
\end{tabular}
\caption{ROC AUC results using input window of length 20. Graph-based methods (NVG and HVG) are compared to the Random Permutation baseline for Real, Synthetic, and Mixed datasets.}
\label{tab:graph_auc_window_20}
\end{table}

The ROC AUC obtained for models trained on real data remains consistently high across both ticker sets (255 and 160). All models show a notable drop in ROC AUC when trained solely on synthetic data, which is expected as synthetic data may not capture all real-world nuances. However, the degree of this drop varies significantly across methods. Models trained on mixed datasets (real and synthetic) tend to perform better than those trained on synthetic data alone, suggesting that blending real data with generated data can help mitigate realism gaps.

Looking at the result for TimeGAN-DS and TimeGAN-SimDS, we observe that the ROC AUC for synthetic data for TimeGAN-SimDS is 0.59581, slightly lower than TimeGAN-DS's 0.60665. ROC AUC for mixed datasets for TimeGAN-SimDS is 0.80691, slightly higher than TimeGAN-DS's 0.80136. This indicates that downsampling by similarity slightly improved generalization when mixed with real data, but not when used on synthetic-only data.

If we compare TimeGAN-DS to TimeGAN-SimDS results for 160 tickers, we see that ROC AUC for synthetic data improves from 0.59583 (TimeGAN-DS) to 0.61375 (TimeGAN-SimDS). ROC AUC for mixed datasets, however, decreases from 0.75867 (TimeGAN) to 0.73463 (TimeGAN-DS). In this case, similarity-based downsampling improved synthetic-only performance, but slightly hurt the mixed-data performance. Comparing DiffusionTS-DS and DiffusionTS-SimDS we observe a consistent advantage for the similarity-based approach. AUC scores improve both on purely synthetic (by +0.83\%) and mixed datasets (by +2.1\%). 

Graph-based methods (NVG, HVG, NVMG) maintain high and consistent performance across all datasets. Even on synthetic data, they show moderate ROC AUC drops (e.g., from 0.82366 to ~0.72–0.73), but still outperform deep generative methods.

Among all evaluated methods, NVG and HVG consistently yielded the highest ROC AUC values for the synthetic dataset, with NVG slightly outperforming others at 0.73071. For the mixed dataset, HVG showed the best performance with an ROC AUC of 0.81206, closely followed by NVG and NVMG. This suggests that traditional visibility graph-based representations are highly effective at capturing structural similarities in both real and synthetic data.

The VRP baseline, which disrupts the temporal structure by shuffling the values within each window while preserving their distribution, performs slightly below the graph-based methods across both experiments. On the dataset with 160 tickers, the Random method achieves a ROC AUC of 0.722 on synthetic data and 0.79514 on mixed data. In comparison, NVG reaches 0.73071 (synthetic) and 0.80331 (mixed), while HVG achieves 0.72598 (synthetic) and 0.81206 (mixed). Similarly, on the larger dataset of 255 tickers, the VRP method scores 0.71443 (synthetic) and 0.83661 (mixed), compared to NVG with 0.74520 (synthetic) and 0.84040 (mixed), and HVG with 0.74891 (synthetic) and 0.84095 (mixed).

Although the differences may appear modest in absolute terms (typically within 1–3 percentage points), the graph-based methods consistently outperform the VRP baseline across both datasets and all conditions. This consistency suggests that visibility graph representations are capturing structural dependencies within the time series that the Random method, by design, cannot retain. The advantage becomes more evident in synthetic datasets, where temporal patterns are often weaker or distorted. These results highlight the ability of graph-based methods to extract and preserve informative patterns in time series, contributing to more robust performance, even when the input data is artificially generated.

We conducted additional experiments comparing our graph-based methods (NVG and HVG) to the VRP baseline across two different time window lengths: 20 and 60. In both settings, we used the full dataset of available tickers. For each window, the classification task involves using the first $n-1$
values to predict whether the final value in the window will go up or down. This setup allows us to assess not only the impact of window size on model performance, but also how sensitive the VRP method is to changes in temporal context. By comparing results across both configurations, we aim to better understand how well different methods capture time-dependent structures over short and long horizons.

Tables~\ref{tab:graph_auc_window_20} and~\ref{tab:graph_auc_window_60} present results for two different input window sizes (20 and 60 time steps) across all dataset types. In both settings, our graph-based methods (NVG and HVG) consistently outperform the VRP baseline, with the most notable differences observed on the synthetic datasets. For the shorter window (w=20), NVG achieves an AUC of 0.85495 on synthetic data, compared to 0.82868 for the VRP baseline. As the window size increases to 60, NVG still leads with an AUC of 0.73463, while the VRP method reaches only 0.70947. This pattern reveals that while the VRP baseline shows minimal benefit from longer sequences, our graph-based methods are capable of leveraging the additional temporal context to extract richer structural information. Moreover, the strong and stable performance of NVG and HVG on both real and mixed datasets further demonstrates their robustness across sequence lengths and data types. These findings support the conclusion that visibility graph representations are highly effective in modeling temporal dependencies.

\subsection{Runtime performance}

The runtime measurements for each method were collected by recording the execution time required to process each ticker individually, across a set of 160 tickers. These per-ticker times were then summed to estimate the total time it would take to run each method sequentially. It is important to note, however, that in practice, some of the methods were executed in parallel, meaning the actual wall-clock time was often shorter than the reported cumulative values. The reported times, therefore, reflect the total computational effort assuming a serial execution, which provides a standardized basis for comparison across methods.

An exception to this measurement approach is the Natural Visibility Multigraphs (NVMG) method. Unlike other methods where synthetic time series were generated per ticker, NVMG constructs graphs jointly for all tickers within defined time segments. Consequently, the runtime for NVMG was measured per time segment rather than per ticker, and the total time was computed by summing the durations associated with each segment.

The runtime results represented in Table \ref{tab:runtime_comparison} clearly demonstrate a significant computational advantage of our proposed methods—NVG, HVG, and NVMG—over deep generative models such as TimeGAN, cGAN, and STS. While the deep learning-based approaches often require hours or even days of training time due to their complexity and data requirements, our graph-based methods complete execution in a matter of seconds or minutes. This stark contrast underscores the efficiency and scalability of visibility graph-based approaches, making them highly suitable for large-scale or real-time financial time series analysis.

\begin{table}[htbp]
\centering
\begin{tabular}{lr}
\toprule
\textbf{Method} & \textbf{Time (days hh:mm:ss)} \\
\midrule
NVG         & 0 00:00:39 \\
HVG         & 0 00:00:41 \\
NVMG        & 0 00:43:05 \\
TimeGAN     & 4 15:10:40 \\
cGAN        & 59 21:12:51 \\
STS         & 0 21:10:55 \\
DiffusionTS & 2 23:06:15 \\
VRP (Ablation)      & \textbf{0 00:00:04} \\
\bottomrule
\end{tabular}
\caption{Runtime comparison of methods. The runtime values are reported in days hh:mm:ss format.}
\label{tab:runtime_comparison}
\end{table}

\section{Conclusions}\label{S:CONCLUSIONS}

As financial markets increasingly rely on data-driven and machine learning approaches, the constraints that historical data poses to such approaches, such as data scarcity, privacy constraints, and the inability to capture rare future scenarios, underscore the importance of high-quality synthetic data generation.

In this work we present a set of novel methods for synthetic data generation in the time series domain, which we comprehensively evaluated on financial time series data. The generative methods leverage graph-based time series representations (NVG, HVG, and NVG-based multigraphs). We assessed their performance against state-of-the-art methods on three diffent criteria: (a) realism of values generated, (b) their performance when considering downstream tasks, and (c) their runtime performance. Our experimental results show that the proposed methods consistently ranked best across all of the evaluation criteria. In particular, the t-SNE visualizations show that data generated by our graph-based models exhibits substantial overlap with real data in latent space, indicating high structural fidelity. Among the rest of the models considered in the experiment, the only model that performed moderately well was TimeGAN.

When considering how synthetic data can be useful for downstream tasks, we observed that synthetic data generated by NVG and HVG consistently achieved the highest ROC AUC scores among all methods. Mixing synthetic and real-world data resulted in an improved ROC AUC when comparing the outcomes of the models trained solely on synthetic data, but a degradation of the performance obtained by considering only real-world data. This suggests that while the synthetic data approximates the real-world data, some relevant information is not sufficiently well captured. Visibility-based methods maintained consistent performance across two dataset sizes (160 and 255 tickers), and across multiple evaluation settings (synthetic-only and mixed datasets).

A key advantage of our proposed model is its computational efficiency.  Our methods significantly outperformed deep learning approaches in runtime, completing execution in seconds or minutes compared to hours or days required by state-of-the-art models. This makes our approach highly practical for real-time or large-scale deployment.

Our future work will focus on synthetic data generation for rare event dynamics and enriching them with labels. Synthetic data generation for rare event dynamics could be crucial to increase the robustness of existing machine learning models to specific scenarios, as well as to enhance reaction times when such scenarios begin to develop. Furthermore, we are interested in understanding whether economic periods (e.g., crises, growth, recovery) can be fingerprinted as to allow for a parameterized scenario generation based on them and the methods we have developed.

\begin{acks}
The work is funded through the projects Graph-Massivizer (HE 101093202), CauseFinder (PNRR 760049), and DataPACT (HE 101189771).
\end{acks}

\bibliographystyle{ACM-Reference-Format}
\bibliography{main}

\end{document}